\documentclass[11pt,a4paper]{article}
\usepackage[hyperref]{acl2017}
\usepackage{times}
\usepackage{latexsym}
\usepackage{graphicx}
\usepackage{amssymb}
\usepackage{url}
\usepackage{enumitem} 
\usepackage[normalem]{ulem}  
\usepackage{comment}
\usepackage{rotating}

\newcommand{\ed}[1]{{\color{black} #1}}

\aclfinalcopy 

  \title{WASSA-2017 Shared Task on Emotion Intensity}

\author{\hspace*{-4mm} Saif M. Mohammad\\
\hspace*{-4mm} Information and Communications Technologies\\
	    \hspace*{-4mm} National Research Council Canada\\
        \hspace*{-4mm} Ottawa, Canada\\
	    \hspace*{-4mm} {\tt saif.mohammad@nrc-cnrc.gc.ca} \\\And
  \hspace*{13mm} Felipe Bravo-Marquez \\
  \hspace*{13mm} Department of Computer Science\\ 
  \hspace*{13mm} The University of Waikato\\
\hspace*{13mm} Hamilton, New Zealand\\
  \hspace*{13mm} {\tt felipe.bravo@waikato.ac.nz} \\}

\date{}

\begin{document}
\maketitle
\begin{abstract}

We present  the first shared task on detecting  the intensity of emotion felt by the speaker of a tweet.
We create the first datasets of tweets annotated for anger, fear, joy, and sadness intensities using a technique called best--worst scaling (BWS).
We show that the annotations lead to reliable fine-grained intensity scores (rankings of tweets by intensity).
The data was partitioned into training, development, and test sets for the competition.
Twenty-two teams participated in the shared task, with the best system
obtaining a Pearson correlation of 0.747 with the gold intensity scores.
We summarize the  machine learning setups, resources, and tools used by the participating teams, with a focus on the techniques and resources that are particularly useful  for the task.
The emotion intensity dataset and the shared task are helping improve our understanding
of how we convey more or less intense emotions through language. 


\end{abstract}

\section{Introduction}
\setitemize[0]{leftmargin=*}
\setenumerate[0]{leftmargin=*}

We use language to communicate not only the emotion we are feeling but also the intensity of the emotion. For example, our utterances can convey that we are very angry, slightly sad, absolutely elated, etc. Here, {\it intensity} refers to the degree or amount of an emotion such as anger or sadness.\footnote{Intensity should not be confused with {\it arousal}, which refers to activation--deactivation dimension---the extent to which an emotion is calming or exciting.} Automatically determining the intensity of emotion felt by the speaker has applications in commerce, public health, intelligence gathering, and social welfare.

Twitter has a large and diverse user base which
entails rich 
textual content, including non-standard language such as emoticons, emojis, creatively spelled words ({\it happee}), and hashtagged words ({\it \#luvumom}). 
Tweets are often used to convey one's emotion, opinion, and stance \cite{MohammadSK17}. Thus, automatically detecting emotion intensities in tweets is especially beneficial in applications such as tracking
brand and product perception, tracking support for issues and policies, tracking public health and well-being,
and disaster/crisis management.
Here, for the first time, we present a shared task on automatically detecting intensity of emotion felt by the speaker of a tweet: WASSA-2017 Shared Task on Emotion Intensity.\footnote{\label{STW}\ed{http://saifmohammad.com/WebPages/EmotionIntensity-SharedTask.html}} 

Specifically, given a tweet and an emotion X, the goal is to determine the intensity or degree of emotion X felt by the speaker---a real-valued score between 0 and 1.{\footnote{\ed{Identifying intensity of emotion evoked in the reader, or intensity of emotion felt by an entity mentioned in the tweet, are also useful tasks, and left for future work.}}
A score of 1 means that the speaker feels the highest amount of emotion X.
A score of 0 means that the speaker feels the lowest amount of emotion X.
We first ask human annotators to infer this intensity of emotion from a tweet. Later,
automatic algorithms are tested to determine the extent to which they can replicate human annotations. 
Note that often a tweet does not explicitly state that the speaker is experiencing a particular emotion, but the intensity of emotion felt by the speaker can be inferred nonetheless. Sometimes a tweet is sarcastic or it conveys the emotions of a different entity,  yet the annotators (and automatic algorithms) are to infer, based on the tweet, the extent to which the speaker is likely feeling a particular emotion.  

In order to provide labeled training, development, and test sets for this shared task, we needed to annotate instances for {\it degree} of affect. This is a substantially more difficult undertaking 
than annotating only for the broad affect class: 
respondents are presented with  greater cognitive load and it is particularly hard to ensure consistency
(both across responses by different annotators and within the responses produced by an individual annotator). 
Thus, we used a technique  called
{\it Best--Worst Scaling (BWS)}, also sometimes referred to as {\it Maximum Difference Scaling (MaxDiff)}. 
It is an annotation scheme that  addresses the limitations of traditional rating scales
\cite{Louviere_1991,Louviere2015,maxdiff-naacl2016,KiritchenkoM2017bwsvsrs}.
We used BWS to create the {\it Tweet Emotion Intensity Dataset}, which currently includes four sets of tweets annotated for intensity of anger, fear, joy, and sadness, respectively \cite{MohammadB17starsem}.
These are the first datasets of their kind. 

The competition is organized on a CodaLab website, where participants can upload their submissions, and the leaderboard reports the results.\footnote{https://competitions.codalab.org/competitions/16380}
Twenty-two teams participated in the 2017 iteration of the competition. 
The best performing system, {\it Prayas}, obtained a Pearson correlation of 0.747 with the gold annotations.  Seven teams obtained scores higher than the score obtained by a competitive SVM-based benchmark system (0.66), which we had released at the start of the competition.\footnote{\ed{https://github.com/felipebravom/AffectiveTweets}}
Low-dimensional (dense) distributed representations of words (word embeddings) and sentences (sentence vectors),  along with presence of affect--associated words (derived from affect lexicons) were the most commonly used features.   
Neural network were the most commonly used machine learning architecture. They were used for learning tweet representations as well as for fitting  regression functions. Support vector machines (SVMs) were the second most popular  regression algorithm. Keras and TensorFlow were some of the most widely used libraries. 


The top performing systems used ensembles of  models trained on dense distributed representations of the tweets as well as features drawn from  affect lexicons.
They also made use of a substantially larger number of affect lexicons than systems that did not perform as well.


The emotion intensity dataset and the corresponding shared task are helping improve our understanding of how we convey more or less intense emotions through language. The task also adds a dimensional nature to model of basic emotions, which has traditionally been viewed as categorical (joy or no joy, fear or no fear, etc.). 
On going work with annotations on the same data for valence , arousal, and dominance aims to better understand the relationships between the circumplex model of emotions \cite{russell2003core} and the categorical model of emotions \cite{Ekman92,Plutchik80}.
Even though the 2017 WASSA shared task has concluded, the CodaLab competition website is kept open. Thus new and improved systems can continually be tested. The best results obtained by any system on the 2017 test set can be found on the CodaLab leaderboard.

The rest of the paper is  organized as follows. We begin with related work and a brief background on best--worst scaling (Section 2). In Section 3, we describe how we collected and annotated the tweets for emotion intensity. We also present experiments to determine the quality of the annotations. Section 4 presents details of the shared task setup. In Section 5, we present a competitive SVM-based baseline that uses a number of common text classification features. We describe ablation experiments to determine the impact of different feature types on regression performance. In Section 6, we present the results obtained by the participating systems and summarize their machine learning setups. Finally, we present conclusions and future directions.    
All of the data, annotation questionnaires, evaluation scripts, regression code, and interactive visualizations of the data are made freely available \ed{on the shared task website}.\footnotemark[\getrefnumber{STW}]

\section{Related Work}

\subsection{Emotion Annotation}
Psychologists have argued that  some emotions are more basic than others  \cite{Ekman92,Plutchik80,Parrot01,frijda1988laws}.  However, they disagree on which emotions (and how many) should be classified as basic emotions---some propose 6, some 8, some 20, and so on.
Thus, most efforts in automatic emotion detection have focused 
on a handful of emotions, especially since
manually annotating text for a large number of emotions is arduous. 
\ed{Apart from these categorical models of emotions, certain dimensional models of emotion have also been proposed. The most popular among them, Russell's circumplex model, asserts that all emotions are made up of two core dimensions: valence and arousal \cite{russell2003core}.} 
We created datasets for four emotions that are the most common amongst the many proposals for basic emotions: anger, fear, joy, and sadness. \ed{However, we 
have also begun work on other affect categories, as well as on valence and arousal.}

The vast majority of emotion annotation work provides discrete binary labels to the text instances (joy--nojoy, fear--nofear, and so on) \cite{AlmRS05,AmanS07,brooks2013statistical,NeviarouskayaPI09,Bollen2009}.
The only annotation effort that provided scores for degree of emotion is  by \citet{SemEval2007} as part of one of the SemEval-2007 shared task. 
Annotators were given newspaper headlines and asked to provide scores between 0 and 100 via slide bars in a web interface. 
It is difficult for humans to provide direct scores at such fine granularity.   A common problem is inconsistency in annotations. 
One annotator might assign a score of 79 to a piece of text, whereas another annotator may assign a score of 62 to the same text. It is also common that the same annotator assigns different scores to the same text instance 
at different points in time.
Further, annotators often have a bias towards different parts of the scale, known as {\it scale region bias}.

\subsection{Best--Worst Scaling}

{\it Best--Worst Scaling (BWS)} 
was developed by \citet{Louviere_1991}, building on some ground-breaking research in the 1960’s in mathematical psychology and psychophysics by Anthony A. J. Marley and Duncan Luce.
Annotators are given $n$ items (an $n$-tuple, where $n > 1$ and commonly $n= 4$). They are asked which item is the {\it best} (highest in terms of the property of interest) and which is the {\it worst} (lowest in terms of the property of interest).
When working on $4$-tuples, best--worst annotations are particularly efficient because each best and worst annotation will reveal the order of
five of the six item pairs. For example, for a 4-tuple with items A, B, C, and D, if A is the best, and D is the worst, then A $>$ B, A $>$ C, A $>$ D, B $>$ D, and C $>$ D.

BWS annotations for a set of $4$-tuples can be easily converted into real-valued scores of association 
between the items and the property of interest \cite{Orme_2009,flynn2014}. 
It has been empirically shown that annotations for $2N$ $4$-tuples is sufficient for obtaining reliable scores (where N is the number of items) \cite{Louviere_1991,maxdiff-naacl2016}.\footnote{At its limit, when $n=2$, BWS becomes a {\it paired comparison} \cite{thurstone1927law,david1963method}, but then a much larger set of tuples need to be annotated (closer to $N^2$).}

\ed{\citet{KiritchenkoM2017bwsvsrs} show through empirical experiments that BWS produces more reliable fine-grained scores than scores obtained using rating scales.}
Within the NLP community, Best--Worst Scaling (BWS) has thus far been used only to annotate words: for example, for creating datasets for relational similarity \cite{jurgens-EtAl:2012:STARSEM-SEMEVAL}, word-sense disambiguation \cite{Jurgens2013EmbracingAA}, word--sentiment intensity \cite{NRCJAIR14},
and phrase sentiment composition \cite{maxdiff-naacl2016}. 
 However, we use BWS to annotate whole tweets for intensity of emotion. 


\section{Data}

\citet{MohammadB17starsem} describe how the {\it Tweet Emotion Intensity Dataset} was created. We summarize below the approach used and the key properties of the dataset. Not included in this summary are: (a) experiments showing marked similarities between emotion pairs in terms of how they manifest in language, (b) how training data for one emotion can be used to improve prediction performance for a different emotion,
and (c) an analysis of the impact of hashtag words on emotion intensities. 

For each emotion X, we select 50 to 100 terms that are associated with that emotion at different intensity levels. For example, for the anger dataset, we use the terms: {\it angry, mad, frustrated, annoyed, peeved, irritated, miffed, fury, antagonism,} and so on. For the sadness dataset, we use the terms: {\it sad, devastated, sullen, down, crying, dejected, heartbroken, grief, weeping,} and so on. We will refer to these terms as the {\it query terms}.

\begin{table}[t!]
 \begin{center}
 \small{
 \begin{tabular}{lrl}
 \hline {\bf Emotion} & {\bf Thes.\@ Category} & {\bf Head Word}\\ \hline
 anger &900 &resentment\\
 fear &860 &fear\\
 joy  &836 &cheerfulness\\
 sadness &837 &dejection\\
 \hline
 \end{tabular}
 }
 \caption{\label{tab:cats} {Categories from the Roget's Thesaurus whose words were taken to be the query terms.}
 }
 \end{center}
 \end{table}

We identified the query words for an emotion by first searching 
the {\it Roget's Thesaurus} to find categories that had the focus emotion word (or a close synonym) as the head word.\footnote{The {\it Roget's Thesaurus}
groups words into about 1000 categories, each containing on average about 100 closely related words.
The head word is the word 
that best represents the meaning of the words within that category.} 
The categories chosen for each head word are shown in Table \ref{tab:cats}.
We chose all 
single-word entries listed within these categories to be the query terms for the corresponding focus emotion.\footnote{The full list of query terms is available on request.}
Starting November 22, 2016, and continuing for three weeks, we polled the Twitter API  for tweets that included the query terms.   
We discarded retweets (tweets that start with RT) and
tweets with urls.
We created a subset of the remaining tweets 
by:\\[-18pt]
\begin{itemize}
\item selecting at most 50 tweets per query term.\\[-20pt]
\item selecting at most 1 tweet for every tweeter--query term combination.\\[-18pt]
\end{itemize}
\noindent Thus, the {\it master set of tweets} is 
\ed{not heavily skewed towards some tweeters or query terms.} 

To study the impact of emotion word hashtags on the intensity of the whole tweet, 
\ed{we identified} tweets that had a query term in hashtag form towards the end of the tweet---\ed{specifically, within the trailing portion of the tweet made up solely of hashtagged words.}
\ed{We created copies of these tweets and then 
removed
the hashtag query terms 
from the copies. The updated tweets were then added to the master set.
Finally, our master set \ed{of 7,097 tweets} includes:}\\[-18pt]
\begin{enumerate}
\item {\it Hashtag Query Term Tweets (HQT Tweets)}:\\ \ed{1030 tweets with a query term in the form of a hashtag (\#$<$query term$>$) in the trailing portion of the tweet};\\[-20pt]
\item {\it No Query Term Tweets (NQT Tweets)}:\\ \ed{1030 tweets 
that are copies of `1', but with the hashtagged query term removed;} 
\item  {\it Query Term Tweets (QT Tweets)}:\\ \ed{5037 tweets that include:\\ a. tweets that contain a query term in the form of a word (no \#$<$query term$>$) \\ b. tweets with a query term in hashtag form followed by at least one non-hashtag word.} 
\end{enumerate}
The master set of tweets was then manually annotated 
for intensity of emotion. Table \ref{tab:tdt} shows a breakdown by emotion.

\subsection{Annotating with Best--Worst Scaling}

We followed the procedure described in \citet{maxdiff-naacl2016} to obtain BWS annotations. 
For each emotion, the annotators were presented with four tweets at a time (4-tuples) and asked to select the speakers of the tweets with the highest and lowest emotion intensity.
$2\times N$ (where $N$ is the number of tweets in the emotion set) distinct 4-tuples were randomly generated in such a manner that
each item is seen in eight different 4-tuples, 
and no pair of items occurs in more than one 4-tuple.
\ed{We refer to this as {\it random maximum-diversity selection (RMDS)}. RMDS maximizes the number of unique items that  each  item  co-occurs with  in the 4-tuples. After BWS annotations, this in turn leads to direct comparative ranking information for the maximum number of pairs of items.\footnote{\ed{In combinatorial mathematics, {\it balanced incomplete block design} refers to creating blocks  (or tuples) of a handful items from a set of $N$ items such that each item occurs in the same number of blocks (say $x$) and each pair of distinct items occurs in the same number of blocks (say $y$), where $x$ and $y$ are integers $ge$ 1 \cite{yates1936incomplete}. The set of tuples we create have similar properties, except that since we create only $2N$ tuples, pairs of distinct items either never occur together in a 4-tuple or they occur in exactly one 4-tuple.}}

It is desirable for an item to occur in sets of 4-tuples such that the the maximum intensities in those 4-tuples are spread across the range from low intensity to high intensity, as then the proportion of times an item is chosen as the best is indicative of its intensity score. Similarly, it is desirable for an item to occur in sets of 4-tuples such that the minimum intensities are spread from low to high intensity. However, since the intensities of items are not known before the annotations, RMDS is used.}

Every 4-tuple was annotated by three independent annotators.\footnote{\citet{maxdiff-naacl2016} showed that using just three annotations per 4-tuple produces highly reliable results. Note that since each tweet is seen in eight different 4-tuples, we obtain $8 \times 3 = 24$ judgments over each tweet.} 
The questionnaires used 
were developed 
through internal discussions and pilot annotations. (See the Appendix (8.1) for a sample questionnaire. All questionnaires are also available on the task website.)

The 4-tuples of tweets were uploaded on the crowdsourcing platform, CrowdFlower.
About 5\% of the data was annotated internally beforehand (by the authors). These questions are referred to as gold questions. 
The gold questions are interspersed with other questions.
If one gets a gold question wrong, they are immediately notified of it. If one's accuracy on the gold questions falls below 70\%, they are refused further annotation, and all of their annotations are discarded. This serves as a mechanism to avoid malicious  annotations.\footnote{\ed{In case more than one item can be reasonably chosen as the best (or worst) item, then more than one acceptable gold answers are provided. The goal with the gold annotations is to identify clearly poor or malicious annotators. In case where two items are close in intensity, we want the crowd of annotators to indicate, through their BWS annotations, the relative ranking of the items.}}

The BWS responses were translated into 
scores 
by a simple calculation \cite{Orme_2009,flynn2014}: For
each item $t$, the score is the percentage of times the $t$ was chosen as having the most intensity minus the percentage of times $t$ was chosen as having the least intensity.\footnote{\ed{\citet{maxdiff-naacl2016} provide code for generating tuples from items using RMDS, as well as code for generating scores from BWS annotations: http://saifmohammad.com/WebPages/BestWorst.html}}
\begin{equation}
{\it intensity} (t) = \%{\it most}(t) - \%{\it least} (t)
\end{equation}
Since intensity  of emotion is a unipolar scale, we linearly transformed the the $-100$ to $100$ scores to scores in the range 0 to 1.



\subsection{Reliability of Annotations}

A useful measure of quality is reproducibility of the end result---if repeated independent manual annotations from multiple respondents result in similar intensity rankings (and scores), then one can be confident that the scores capture the true emotion intensities. 
To assess this reproducibility, we calculate average {\it split-half reliability (SHR)}, \ed{a commonly used approach to determine consistency
\ed{\cite{kuder1937theory,cronbach1946case}}. 
The intuition behind SHR is as follows.}   
All 
annotations for an item (in our case, tuples) are randomly split into two halves. Two sets of scores are produced independently from the two halves. 
Then the correlation between the two sets of scores is calculated. If the annotations are of good quality, then the correlation between the two halves will be high.

\ed{Since each tuple in this dataset was annotated by three annotators (odd number), we  calculate SHR by randomly placing one or two annotations per tuple in one bin and the remaining (two or one) annotations for the tuple in another bin. Then two sets of intensity scores (and rankings) are calculated  from the  annotations in each of the two bins. The process is repeated 100 times and the correlations across the  two sets of rankings and intensity scores are averaged.
Table \ref{tab:shr} shows the split-half reliabilities for the anger, fear, joy, and sadness tweets in the {\it Tweet Emotion Intensity Dataset}.\footnote{\ed{Past work has found the SHR for sentiment intensity annotations for words, with 8 annotations per tuple, to be 0.98 \cite{NRCJAIR14}. In contrast, here SHR is calculated from 3 annotations, for emotions, and from whole sentences. SHR determined from a smaller number of annotations and on more complex annotation tasks are expected to be lower.}}  Observe that for fear, joy, and sadness datasets, both the Pearson correlations and the Spearman rank correlations lie between 0.84 and 0.88, indicating a high degree of reproducibility. However, the correlations are slightly lower for anger indicating that it is relative more difficult to ascertain the degrees of anger of speakers from their tweets.
Note that SHR indicates the quality of annotations obtained when using only half the number of annotations. The correlations obtained when repeating the experiment  with three annotations for each 4-tuple is expected to be even higher. 
Thus the numbers shown in Table \ref{tab:shr} are a lower bound on the quality of annotations obtained with three annotations per 4-tuple.
}

\begin{table}[t!]
\begin{center}
{\small
\begin{tabular}{lrrlrrlrrlr}
\hline
\bf Emotion		&\bf Spearman	&\bf Pearson\\\hline
anger		&0.779		&0.797\\
fear		&0.845		&0.850\\
joy			&0.881		&0.882\\
sadness		&0.847		&0.847\\
\hline
\end{tabular}
\caption{\label{tab:shr} {\ed{Split-half reliabilities (as measured by Pearson correlation and Spearman rank correlation) for the anger, fear, joy, and sadness tweets in the Tweet Emotion Intensity Dataset.}}}
}
\end{center}
\vspace*{-3mm}
\end{table}

\section{Task Setup}

\subsection{The Task}
Given a tweet and an emotion X, automatic systems have to determine the intensity or degree of emotion X felt by the speaker---a real-valued score between 0 and 1. 
A score of 1 means that the speaker feels the highest amount of emotion X.
A score of 0 means that the speaker feels the lowest amount of emotion X. 
The competition is organized on a CodaLab website, where participants can upload their submissions, and the leaderboard reports the results.\footnote{https://competitions.codalab.org/competitions/16380}

\begin{table}[t!]
\begin{center}
{\small
\begin{tabular}{lrrrr}
\hline {\bf Emotion} & {\bf Train} & {\bf Dev.} &{\bf Test} &{\bf All}\\ \hline
anger &857 &84 &760 &1701\\
fear &1147 &110 &995 &2252\\
joy  &823 &74 &714 &1611\\
sadness &786 &74  &673 &1533\\ \hline
{\bf All}  &3613 &342 &3142 &7097\\
\hline
\end{tabular}
}
\caption{\label{tab:tdt} {The number of instances in the Tweet Emotion Intensity dataset.}
}
\vspace*{-2mm}
\end{center}
\end{table}

\subsection{Training, development, and test sets}

The {\it Tweet Emotion Intensity Dataset} is partitioned
into training, development, and test sets for machine learning experiments (see Table \ref{tab:tdt}). 
For each emotion, we chose to include about 50\% of the tweets in the training set, about 5\% in the development set, and about 45\% in the test set. Further, we ensured that an No-Query-Term (NQT) tweet is in the same partition as the Hashtag-Query-Term (HQT) tweet it was created from.

The training and development sets were made available more than two months before the two-week official evaluation period.  Participants were told that the development set could be used to tune one’s system and also to test making a submission on CodaLab.
Gold intensity scores for the development set were released two weeks before the evaluation period, and participants were free to train their systems on the combined training and development sets, and apply this model to the test set. 
The test set was released at the start of the evaluation period.

\subsection{Resources}

Participants were free to use lists of manually created and/or automatically generated word--emotion and word--sentiment association lexicons.\footnote{A large number of sentiment and emotion lexicons created at NRC are available here: http://saifmohammad.com/WebPages/lexicons.html}
Participants were free to build a system from scratch or use any available software packages and resources, as long as they are not against the spirit of fair competition. In order to assist testing of ideas, we also provided a baseline Weka system for determining emotion intensity, that participants can build on directly or use to determine the usefulness of different features.\footnote{https://github.com/felipebravom/AffectiveTweets} 
We describe the baseline system in the next section.

\subsection{Official Submission to the Shared Task}

System submissions were required to have the same format as used in the training and test sets. Each line in the file should include:\\
{\tt id}[tab]{\tt tweet}[tab]{\tt emotion}[tab]{\tt score }

Each team was allowed to make as many as ten submissions during the evaluation period. However, they were told in advance that only the final submission would be considered as the official submission to the competition.

Once the evaluation period concluded, we released the gold labels and participants were able to determine results on various system variants that they may have developed. We encouraged participants to report results on all of their systems (or system variants) in the system-description paper that they write. However, they were asked to clearly indicate the result of their official submission. 

During the evaluation period, the CodaLab leaderboard was hidden from participants---so they were unable see the results of their submissions on the test set until the leaderboard was subsequently made public.
Participants were, however, able to immediately see any warnings or errors that their submission may have triggered.

\subsection{Evaluation} 

For each emotion, systems were evaluated by calculating the Pearson Correlation Coefficient of the system predictions with the gold ratings. Pearson coefficient, which  measures linear correlations between two variables, produces scores from -1 (perfectly inversely correlated) to 1 (perfectly correlated).
A score of 0 indicates no correlation.
The correlation scores across all four emotions was averaged to determine the bottom-line competition metric by which the submissions were ranked.

In addition to the bottom-line competition metric described above, the following additional metrics were also provided:
\begin{itemize}
\item Spearman Rank Coefficient of the submission with the gold scores of the test data.\\
Motivation: Spearman Rank Coefficient considers only how similar the two sets of ranking are. The differences in scores between adjacently ranked instance pairs is ignored. On the one hand this has been argued to alleviate some biases in Pearson, but on the other hand it can ignore relevant information.
\item Correlation scores (Pearson and Spearman) over a subset of the testset formed by taking instances with gold intensity scores $\ge 0.5$.\\
Motivation: In some applications, only those instances that are moderately or strongly emotional are relevant. Here it may be much more important for a system to correctly determine emotion intensities of instances in the higher range of the scale as compared to correctly determine emotion intensities in the lower range of the scale.
\end{itemize}
\noindent Results with Spearman rank coefficient were largely inline with those obtained using Pearson coefficient, and so in the rest of the paper we report only the latter.
However, the  CodaLab leaderboard and the official results posted on the task website show both metrics.
The official evaluation script (which calculates correlations using both metrics and also acts as a format checker) was made available along with the training and development data well in advance. Participants were able to use it to monitor progress of their system by cross-validation on the training set or testing on the development set. The script was also uploaded on the CodaLab competition website so that the system evaluates submissions automatically and updates the leaderboard. 


\begin{table*}[t]
\begin{center}
\resizebox{0.85\textwidth}{!}{
\begin{tabular}{lllll}
\hline
 &\bf Twitter &\bf Annotation &\bf  Scope &\bf Label \\ \hline
AFINN \cite{nielsen2011new} & Yes & Manual & Sentiment & Numeric \\ 
BingLiu \cite{Liu2004} & No & Manual & Sentiment & Nominal \\ 
MPQA \cite{Wilson05} & No & Manual & Sentiment & Nominal \\ 
NRC Affect Intensity Lexicon (NRC-Aff-Int) \cite{mohammad2017word} & Yes & Manual & Emotions & Numeric\\
NRC Word-Emotion Assn.\@ Lexicon (NRC-EmoLex) \cite{MohammadT13} & No & Manual & Emotions & Nominal\\ 
NRC10 Expanded (NRC10E) \cite{bravo2016determining} & Yes & Automatic & Emotions & Numeric \\ 
NRC Hashtag Emotion Association Lexicon (NRC-Hash-Emo) & Yes & Automatic & Emotions & Numeric \\ 
$\;\;\;\;\;$ \cite{mohammad:2012:STARSEM-SEMEVAL,COIN:COIN12024} &\\
NRC Hashtag Sentiment Lexicon (NRC-Hash-Sent)  \cite{MohammadSemEval2013} & Yes & Automatic & Sentiment & Numeric \\ 
Sentiment140 \cite{MohammadSemEval2013} & Yes & Automatic & Sentiment & Numeric \\ 
SentiWordNet \cite{Esuli06} & No & Automatic & Sentiment & Numeric \\ 
SentiStrength \cite{ThelwallBP12} & Yes & Manual & Sentiment & Numeric \\ \hline
\end{tabular}}
\end{center}
\vspace*{-2mm}
\caption{Affect lexicons used in our experiments.}
\label{tab:lex_prop}
\end{table*}

\section{Baseline System for Automatically Determining Tweet Emotion Intensity}

\subsection{System}
\ed{We implemented a package called AffectiveTweets \cite{MohammadB17starsem} for the Weka machine learning workbench \cite{Wekapaper}. It provides a collection of filters for extracting 
 features from tweets for sentiment classification and other related tasks. These include features used in \citet{NRCJAIR14} and \citet{MohammadSK17}.\footnote{\citet{NRCJAIR14} describes the NRC-Canada system which ranked first in three sentiment shared tasks: SemEval-2013 Task 2, SemEval-2014 Task 9, and SemEval-2014 Task 4. \citet{MohammadSK17} describes a stance-detection system that outperformed submissions from all 19 teams that participated in SemEval-2016 Task 6.}
 We use the AffectiveTweets package for calculating feature vectors from our emotion-intensity-labeled tweets and train Weka regression models on this transformed data.}
The regression model used is an $L_{2}$-regularized $L_{2}$-loss SVM regression model with the  regularization parameter $C$ set to 1, implemented in LIBLINEAR\footnote{{http://www.csie.ntu.edu.tw/$\sim$cjlin/liblinear/}}. 
The system uses the following features:\footnote{See Appendix (A.3) for further implementation details.}

\noindent {\it a. Word N-grams (WN)}: presence or absence of word n-grams from $n=1$ to $n=4$. \\[-15pt]

\noindent {\it b. Character N-grams (CN)}: presence or absence of character n-grams from $n=3$ to $n=5$. 
\\[-15pt]

\noindent {\it c. Word Embeddings (WE)}: an average of the word embeddings of all the words in a tweet. We calculate individual word embeddings using the negative sampling skip-gram model implemented in \textit{Word2Vec} \cite{mikolov2013efficient}. 
Word vectors are trained from ten million English tweets taken from the Edinburgh Twitter Corpus \cite{Petrovic2010}. We set {\it Word2Vec} parameters: window size: 5; number of dimensions: 400.\footnote{\ed{Optimized for the task of word--emotion classification on an independent dataset \cite{bravo2016determining}.}}\\[-15pt]

\noindent {\it d. Affect Lexicons (L)}: we 
use the lexicons shown in Table \ref{tab:lex_prop} by aggregating the information for all the words in a tweet. 
If the lexicon provides nominal association labels (e.g, positive, anger, etc.), then the number of words in the tweet matching each class are counted. 
If the lexicon provides numerical scores, the individual scores for each class are summed. 
and whether the affective associations provided are nominal or numeric. 

\subsection{Experiments}
\label{sec:exp1}

We  \ed{developed the baseline system by learning models from each of the {\it Tweet Emotion Intensity Dataset} training sets and
applying them to the corresponding development sets. Once the system parameters were frozen, the system learned new models from the combined training and development corpora. This model was applied to the test sets.}
Table  \ref{tab:reg_res_full} shows the results obtained on the test sets using
various features, individually and in combination. The last column `avg.' shows the
macro-average of the  correlations for all of the emotions.  

Using just character or just word n-grams leads to results around 0.48, suggesting that they are reasonably good indicators of emotion intensity by themselves. (Guessing the  intensity scores at random between 0 and 1 is expected to get correlations close to 0.)
Word embeddings produces 
statistically significant 
improvement over the ngrams (avg.\@ r = 0.55).\footnote{We used the
Wilcoxon signed-rank test at 0.05 significance level calculated from ten random partitions of the data,
for all the significance tests reported in this paper.}
Using features drawn from affect lexicons produces results ranging from  
avg.\@ r = 0.19  with SentiWordNet to  avg.\@ r =  0.53 with NRC-Hash-Emo.  
Combining all the lexicons  leads to 
statistically significant 
improvement over individual lexicons (avg.\@ r = 0.63). 
Combining the different kinds of features  leads to even higher scores, with the best overall result obtained using  
word embedding and lexicon features (avg.\@ r = 0.66).\footnote{The increase from 0.63 to 0.66 is statistically significant.}
The feature space formed by all the lexicons together 
is the strongest single feature category. The results also show that some features such as character ngrams are redundant in the presence of certain other features.

Among the lexicons, NRC-Hash-Emo is the most predictive single lexicon.  Lexicons that include Twitter-specific entries, 
lexicons that include intensity scores, and lexicons that label emotions and not just sentiment, tend to be more  predictive on this task--dataset combination. 
\ed{NRC-Aff-Int has real-valued fine-grained word--emotion association scores for all the words in NRC-EmoLex that were marked as being associated with anger, fear, joy, and sadness.\footnote{\ed{\scalebox{0.95}{http://saifmohammad.com/WebPages/AffectIntensity.htm}}}
Improvement in scores obtained  using NRC-Aff-Int  over the scores obtained using NRC-EmoLex also show that using fine intensity scores of word-emotion association are beneficial for tweet-level emotion intensity detection.}
The correlations for anger, fear, and joy 
 are similar 
(around 0.65), but the correlation for sadness is markedly higher (0.71).  We can observe from 
Table \ref{tab:reg_res_full}  
that this  boost in performance for sadness is to some extent due to word embeddings, but is more so due to  lexicon features, 
especially those from  SentiStrength. 
SentiStrength focuses solely on positive and negative classes, but provides numeric scores for each. 

To assess performance in the moderate-to-high range of the intensity scale, we calculated correlation scores over a subset of the test data formed by taking only those instances with gold emotion intensity scores $\geq 0.5$. 
The last row in Table~\ref{tab:reg_res_full} shows the results.  We observe that the correlation scores are in general lower here in the 0.5 to 1 range of intensity scores than in the experiments over the full intensity range.  This is simply because this is a harder task as now the systems do not benefit by making coarse distinctions over whether a tweet is  in the lower range or in the higher range.

\begin{table}[t!]
\resizebox{0.48\textwidth}{!}{
\begin{tabular}{lrrrrr}
\hline
 & \multicolumn{5}{c}{\bf Pearson correlation r}\\
 \multicolumn{2}{r}{\bf anger} &\bf fear &\bf joy & \bf sad. &\bf avg.\\ \hline
{\it Individual feature sets}\\
$\;\;\;$ word ngrams (WN) 						& 0.42 & 0.49 & 0.52 & 0.49 & 0.48 \\ 
$\;\;\;$ char.\@ ngrams (CN) 						& 0.50 & 0.48 & 0.45 & 0.49 & 0.48 \\ 
$\;\;\;$ word embeds.\@ (WE) 						& 0.48 & 0.54 & 0.57 & 0.60 & 0.55 \\ 
$\;\;\;$ all lexicons (L) 						&\bf 0.62 &\bf 0.60 &\bf 0.60 &\bf 0.68 &\bf 0.63 \\ 
$\;\;\;$ \it Individual Lexicons &\\
$\;\;\;\;\;\;$ AFINN 					& 0.48 & 0.27 & 0.40 & 0.28 & 0.36 \\ 
$\;\;\;\;\;\;$ BingLiu 				& 0.33 & 0.31 & 0.37 & 0.23 & 0.31 \\ 
$\;\;\;\;\;\;$ MPQA 					& 0.18 & 0.20 & 0.28 & 0.12 & 0.20 \\ 
$\;\;\;\;\;\;$ NRC-Aff-Int 					& 0.24 & 0.28 & 0.37 & 0.32 & 0.30 \\
$\;\;\;\;\;\;$ NRC-EmoLex 					& 0.18 & 0.26 & 0.36 & 0.23 & 0.26 \\ 
$\;\;\;\;\;\;$ NRC10E 					& 0.35 & 0.34 & 0.43 & 0.37 & 0.37 \\ 
$\;\;\;\;\;\;$ NRC-Hash-Emo 			&\bf 0.55 &\bf 0.55 &\bf 0.46 & 0.54 &\bf 0.53 \\ 
$\;\;\;\;\;\;$ NRC-Hash-Sent 			& 0.33 & 0.24 & 0.41 & 0.39 & 0.34 \\ 
$\;\;\;\;\;\;$ Sentiment140 			& 0.33 & 0.41 & 0.40 & 0.48 & 0.41 \\ 
$\;\;\;\;\;\;$ SentiWordNet 			& 0.14 & 0.19 & 0.26 & 0.16 & 0.19 \\ 
$\;\;\;\;\;\;\;$SentiStrength 			& 0.43 & 0.34 &\bf 0.46 &\bf 0.61 & 0.46 \\
{\it Combinations} &\\
$\;\;\;$ WN + CN + WE 			& 0.50 & 0.48 & 0.45 & 0.49 & 0.48 \\ 
$\;\;\;$ WN + CN + L 			& 0.61 & 0.61 & 0.61 & 0.63 & 0.61 \\ 
$\;\;\;$ WE + L 					& \textbf{0.64} & 0.63 & \textbf{0.65} & \textbf{0.71} & \textbf{0.66}\\
$\;\;\;\;$WN + WE + L 			& 0.63 & \textbf{0.65} & \textbf{0.65} & 0.65 & 0.65 \\ 
$\;\;\;$ CN + WE + L 			& 0.61 & 0.61 & 0.62  & 0.63 & 0.62 \\ 
$\;\;\;$ WN + CN + WE + L 		& 0.61 & 0.61 & 0.61 & 0.63 & 0.62 \\
\\
\multicolumn{6}{l}{\it Over the subset of test set where intensity $\ge 0.5$} \\
$\;\;\;$ WN + WE + L 			& 0.51 & 0.51 & 0.40 & 0.49 & 0.47 \\ 
\hline
\end{tabular}
}
\caption{Pearson correlations (r) of emotion intensity predictions with gold scores. Best results for each column are shown in bold: highest score by a 
feature set, highest score using a single lexicon, and highest score using feature set combinations.}
\label{tab:reg_res_full}
 \vspace*{-2mm}
\end{table}

\section{Official System Submissions to the Shared Task}

Twenty-two teams made submissions to the shared task. In the subsections below we present the results and summarize the approaches and resources used by the participating systems.

\begin{table*}[htbp]
\center
{\small
\begin{tabular}{lrrrrr}
\hline
\bf Team Name & \textbf{r avg. (rank)} &  \textbf{r fear (rank)} &\textbf{r joy (rank)} & \textbf{r sadness (rank)}  &\bf r anger (rank) \\ \hline
1. Prayas & 0.747 (1) & 0.732 (1) & 0.762 (1) & 0.732 (1) & 0.765 (2) \\ 
2. IMS & 0.722 (2) & 0.705 (2) & 0.726 (2) & 0.690 (4) & 0.767 (1) \\ 
3. SeerNet & 0.708 (3) & 0.676 (4) & 0.698 (6) & 0.715 (2) & 0.745 (3) \\ 
4. UWaterloo & 0.685 (4) & 0.643 (8) & 0.699 (5) & 0.693 (3) & 0.703 (7) \\ 
5. IITP & 0.682 (5) & 0.649 (7) & 0.713 (4) & 0.657 (7) & 0.709 (5) \\ 
6. YZU NLP & 0.677 (6) & 0.666 (5) & 0.677 (8) & 0.658 (6) & 0.709 (5) \\ 
7. YNU-HPCC & 0.671 (7) & 0.661 (6) & 0.697 (7) & 0.599 (9) & 0.729 (4) \\ 
8. TextMining & 0.649 (8) & 0.604 (10) & 0.663 (9) & 0.660 (5) & 0.668 (10) \\ 
9. XRCE & 0.638 (9) & 0.629 (9) & 0.657 (10) & 0.594 (10) & 0.672 (9) \\ 
10. LIPN & 0.619 (10) & 0.58 (11) & 0.639 (11) & 0.583 (11) & 0.676 (8) \\ 
11. DMGroup & 0.571 (11) & 0.55 (12) & 0.576 (12) & 0.556 (12) & 0.603 (11) \\ 
12. Code Wizards & 0.527 (12) & 0.465 (16) & 0.534 (15) & 0.532 (14) & 0.578 (13) \\ 
13. Todai & 0.522 (13) & 0.470 (15) & 0.561 (13) & 0.537 (13) & 0.520 (16) \\ 
14. SGNLP & 0.494 (14) & 0.486 (14) & 0.512 (16) & 0.429 (18) & 0.550 (14) \\ 
15. NUIG & 0.494 (14) & 0.680 (3) & 0.717 (3) & 0.625 (8) & -0.047 (21) \\ 
16. PLN PUCRS & 0.483 (16) & 0.508 (13) & 0.460 (19) & 0.425 (19) & 0.541 (15) \\ 
17. H.Niemtsov & 0.468 (17) & 0.412 (17) & 0.511 (17) & 0.437 (17) & 0.513 (17) \\ 
18. Tecnolengua & 0.442 (18) & 0.373 (18) & 0.488 (18) & 0.439 (16) & 0.469 (18) \\ 
19. GradAscent & 0.426 (19) & 0.356 (19) & 0.543 (14) & 0.226 (20) & 0.579 (12) \\ 
20. SHEF/CNN & 0.291 (20) & 0.277 (20) & 0.109 (20) & 0.517 (15) & 0.259 (19) \\ 
21. deepCybErNet & 0.076 (21) & 0.176 (21) & 0.023 (21) & -0.019 (21) & 0.124 (20) \\
{\it Late submission}\\
$*$ SiTAKA  & 0.631 & 0.626 & 0.619 & 0.593 & 0.685 \\ \hline
\end{tabular}}
\caption{Official Competition Metric: Pearson correlations (r) and ranks (in brackets) obtained by the systems on the full test sets. The bottom-line competition metric, `r avg.', is the average of Pearson correlations obtained for each of the four emotions.}
\label{tab:leaderboard_full}
\end{table*}

\begin{table*}[htbp]
\center
{\small
\begin{tabular}{lrrrrr}
\hline
\bf Team Name & \textbf{r avg. (rank)} &  \textbf{r fear (rank)} &\textbf{r joy (rank)} & \textbf{r sadness (rank)}  &\bf r anger (rank)  \\ \hline
1. Prayas & 0.571 (1) & 0.605 (1) & 0.621 (1) & 0.500 (2) & 0.557 (2) \\ 
3. SeerNet & 0.547 (2) & 0.529 (5) & 0.551 (7) & 0.551 (1) & 0.556 (3) \\ 
4. UWaterloo & 0.520 (3) & 0.499 (9) & 0.562 (4) & 0.480 (3) & 0.538 (4) \\ 
6. YZU NLP & 0.516 (4) & 0.544 (3) & 0.552 (5) & 0.471 (5) & 0.495 (7) \\ 
2. IMS & 0.514 (5) & 0.519 (7) & 0.552 (5) & 0.415 (7) & 0.570 (1) \\ 
5. IITP & 0.505 (6) & 0.525 (6) & 0.575 (2) & 0.406 (8) & 0.513 (6) \\ 
7. YNU-HPCC & 0.500 (7) & 0.530 (4) & 0.540 (8) & 0.406 (8) & 0.526 (5) \\ 
8. TextMining & 0.486 (8) & 0.480 (10) & 0.513 (9) & 0.472 (4) & 0.477 (9) \\ 
9. XRCE & 0.450 (9) & 0.506 (8) & 0.507 (10) & 0.357 (14) & 0.430 (12) \\ 
10. LIPN & 0.446 (10) & 0.435 (12) & 0.496 (11) & 0.366 (12) & 0.489 (8) \\ 
11. DMGroup & 0.432 (11) & 0.456 (11) & 0.483 (13) & 0.329 (16) & 0.462 (10) \\ 
15. NUIG & 0.390 (12) & 0.567 (2) & 0.566 (3) & 0.426 (6) & 0.003 (21) \\ 
13. Todai & 0.387 (13) & 0.350 (15) & 0.484 (12) & 0.362 (13) & 0.351 (17) \\ 
12. Code Wizards & 0.380 (14) & 0.344 (16) & 0.422 (16) & 0.318 (17) & 0.437 (11) \\ 
14. SGNLP & 0.373 (15) & 0.386 (13) & 0.390 (17) & 0.330 (15) & 0.387 (16) \\ 
19. GradAscent & 0.367 (16) & 0.245 (19) & 0.457 (14) & 0.376 (11) & 0.392 (15) \\ 
17. H.Niemtsov & 0.347 (17) & 0.275 (17) & 0.441 (15) & 0.242 (18) & 0.428 (13) \\ 
16. PLN PUCRS & 0.313 (18) & 0.361 (14) & 0.315 (18) & 0.155 (19) & 0.424 (14) \\ 
20. SHEF/CNN & 0.220 (19) & 0.188 (21) & 0.095 (20) & 0.396 (10) & 0.202 (20) \\ 
18. Tecnolengua & 0.209 (20) & 0.247 (18) & 0.224 (19) & 0.061 (20) & 0.305 (18) \\ 
21. deepCybErNet & 0.140 (21) & 0.190 (20) & 0.077 (21) & 0.057 (21) & 0.235 (19) \\ 
{\it Late submission}\\
$*$ SiTAKA & 0.484 & 0.496 & 0.46 & 0.465 & 0.513 \\ \hline
\end{tabular}}
\caption{Pearson correlations (r) and ranks (in brackets) obtained by the systems on a subset of the test set where gold scores $\geq 0.5$}
\label{tab:leaderboard_hr}
\end{table*}

\subsection{Results}

Table \ref{tab:leaderboard_full} shows the  Pearson correlations (r) and ranks (in brackets) obtained by the systems on the full test sets. The bottom-line competition metric, `r avg.', is the average of Pearson correlations obtained for each of the four emotions.  (The task website shows  Spearman rank coefficient as well.  Those scores are close in value to the Pearson correlations, and most teams rank the same by either metric.)
 The top ranking system, {\it Prayas}, obtained an {\it r avg.} of  0.747. 
 It obtains slightly better correlations for joy and anger (around 0.76) than for fear and sadness  (around 0.73).
 {\it IMS}, which ranked second overall, obtained slightly higher correlation on anger, but lower scores than {\it Prayas} on the other emotions.
The top 12 teams all obtain their best correlation on anger as opposed to any of the other three emotions. They obtain lowest correlations on fear and sadness.
Seven  teams obtained scores higher than that obtained by the publicly available benchmark system (r avg.\@ = 0.66).  

Table \ref{tab:leaderboard_hr} shows the  Pearson correlations (r) and ranks (in brackets) obtained by the systems on those instances in the test set with intensity scores $\ge 0.5$.  {\it Prayas} obtains the best results here too with r avg.\@ =  0.571. {\it SeerNet}, which ranked third on the full test set, ranks second on this subset. As found in the baseline results, system results on this subset overall are lower than than on the full test set.  Most  systems  perform best on the joy data and worst on the sadness data.


\subsection{Machine Learning Setups}
Systems followed a supervised learning approach in which tweets were mapped into feature vectors 
that were then used for training regression models.

Features were drawn both from the training data as well as from external resources such as large tweet corpora and affect lexicons. Table \ref{tab:features} lists the feature types (resources) used by the teams.  (To save space, team names are abbreviated to just their rank on the full test set (as shown in Table 6).)
Commonly used features included word embeddings and sentence representations learned using neural networks (sentence embeddings). 
Some of the word embeddings models used were Glove (SeerNet, UWaterloo, YZU NLP), Word2Vec (SeerNet), and Word Vector Emoji Vectors (SeerNet).
The models used for learning sentence embeddings included LSTM  (Prayas, IITP), CNN (SGNLP), LSTM--CNN combinations (IMS, YMU-HPCC), bi-directional versions (YZU NLP), and augmented LSTMs models with attention layers (Todai). 
High-dimensional sparse representations such as word n-grams or character n-grams were rarely used.  
Affect lexicons were also  widely used, especially by the top eight teams.
Some teams built their own affect lexicons from additional data (IMS, XRCE). 

The regression algorithms applied to the feature vectors included SVM regression or SVR (IITP, Code Wizards, NUIG, H.Niemstov), Neural Networks (Todai, YZU NLP, SGNLP),  Random Forest (IMS, SeerNet, XRCE), Gradient Boosting (UWaterLoo, PLN PUCRS), AdaBoost (SeerNet), and Least Square Regression (UWaterloo).
Table~\ref{tab:regression} provides the full list. 

Some teams followed a popular deep learning trend wherein the feature representation and the prediction model are trained in conjunction. In those systems, the regression algorithm corresponds to the output layer of the neural network (YZU NLP, SGNLP, Todai).

Many libraries and tools were used for implementing the systems. The high-level neural networks API library \textit{Keras} was the most widely used off-the-shelf package. It is written in Python and runs on top of either {\it TensorFlow} or {\it Theano}. \textit{TensorFlow} and \textit{Sci-kit learn} were also popular (also Python libraries).\footnote{TensorFlow provides implementations of a number of machine learning algorithms, including deep learning ones such as CNNs and LSTMs.} Our AffectiveTweets Weka baseline package was used by five participating teams, including the teams that ranked first, second, and third. The full list of tools and libraries used by the 
teams is shown in Table~\ref{tab:tools}.

\begin{table*}[htbp]
\begin{center}
\resizebox{0.9\textwidth}{!}{
\begin{tabular}{lcccccccccccccccccccccc}
\hline
   &\multicolumn{22}{c}{\bf Team}\\
Features	&1 &2 &3 &4 &5 &6 &7 &8 &9 &$*$ &10 &11 &12 &13  &14 &15 &16 &17 &18 &19 &20 &21\\  \hline
N-grams &  &  &  & \checkmark  &  &  &  &  &  &  &  &   & \checkmark &  &  &  &  &  &  &  &  &  \\ 
$\;\;\;$ CN &  &  &  &  &  &  &  &  &  &  &  &  & \checkmark &  &  &  &  &  &  &  &  &  \\ 
$\;\;\;$ WN &  &  &  & \checkmark &  &  &  &  &  &  &  &  & \checkmark &  &  & \checkmark &  &  &  &  &  &  \\ 
Word Embeddings & \checkmark & \checkmark & \checkmark & \checkmark & \checkmark & \checkmark & \checkmark & \checkmark &  & \checkmark &  &  & \checkmark & \checkmark & \checkmark & \checkmark &  &  &  & \checkmark &  &  \\ 
$\;\;\;$ Glove &  &  & \checkmark & \checkmark & \checkmark & \checkmark & \checkmark & \checkmark &  & \checkmark &  &  &  & \checkmark &  & \checkmark &  &  &  & \checkmark &  &  \\ 
$\;\;\;$ Emoji Vectors &  &  & \checkmark & \checkmark &  &  &  &  &  &  &  &  &  &  &  &  &  &  &  &  &  &  \\ 
$\;\;\;$ Word2Vec & \checkmark & \checkmark & \checkmark & \checkmark &  &  &  &  &  &  &  &  &  &  &  &  &  &  &  &  &  &  \\ 
$\;\;\;$ Other &  &  &  &  &  &  &  & \checkmark &  &  &  &  & \checkmark &  & \checkmark &  &  &  &  &  &  &  \\ 
Sentence Embeddings &  &  &  &  &  &  &  &  &  &  &  &  &  &  &  &  &  &  &  &  &  &  \\ 
$\;\;\;$ CNN & \checkmark & \checkmark &  &  &  & \checkmark & \checkmark & \checkmark &  & \checkmark &  &  &  &  & \checkmark &  &  &  &  &  & \checkmark & \checkmark \\ 
$\;\;\;$ LSTM & \checkmark & \checkmark &  &  & \checkmark & \checkmark & \checkmark & \checkmark &  &  &  &  &  & \checkmark &  & \checkmark &  &  &  & \checkmark &  &  \\ 
$\;\;\;$ Other &  &  &  & \checkmark &  &  &  &  &  &  &  &  &  &  &  & \checkmark &  &  &  & \checkmark & \checkmark &  \\ 
Affective Lexicons &  & \checkmark & \checkmark & \checkmark & \checkmark & \checkmark &  & \checkmark & \checkmark & \checkmark &  &  &  & \checkmark &  &  &  & \checkmark & \checkmark & \checkmark &  &  \\ 
$\;\;\;$ AFINN & \checkmark & \checkmark & \checkmark &  & \checkmark &  &  & \checkmark &  &  &  &  &  &  &  &  &  &  &  &  &  &  \\ 
$\;\;\;$ ANEW &  & \checkmark &  &  &  &  &  &  &  &  &  &  &  &  &  &  &  &  &  &  &  &  \\ 
$\;\;\;$ BingLiu & \checkmark & \checkmark & \checkmark &  & \checkmark &  &  & \checkmark & \checkmark &  &  &  &  &  &  &  &  &  &  &  &  &  \\ 
$\;\;\;$ Happy Ratings &  & \checkmark &  &  &  &  &  &  &  &  &  &  &  &  &  &  &  &  &  &  &  &  \\ 
$\;\;\;$ Lingmotif &  &  &  &  &  &  &  &  &  &  &  &  &  &  &  &  &  &  & \checkmark &  &  &  \\ 
$\;\;\;$ LIWC &  &  &  &  &  &  &  &  &  &  &  &  &  &  &  &  & \checkmark &  &  &  &  &  \\ 
$\;\;\;$ MPQA & \checkmark & \checkmark & \checkmark &  & \checkmark &  &  & \checkmark &  &  &  &  &  &  &  &  &  &  &  &  &  &  \\ 
$\;\;\;$ NRC-Aff-Int  & \checkmark &  & \checkmark & \checkmark &  &  &  & \checkmark &  &  &  &  &  &  &  &  &  &  &  &  &  &  \\ 
$\;\;\;$ NRC-EmoLex  & \checkmark & \checkmark & \checkmark & \checkmark & \checkmark &  &  & \checkmark & \checkmark &  &  &  &  &  &  &  &  &  &  &  &  &  \\ 
$\;\;\;$ NRC-Emoticon-Lex & \checkmark &  & \checkmark & \checkmark &  &  &  & \checkmark &  &  &  &  & \checkmark &  &  &  &  &  &  &  &  &  \\
$\;\;\;$ NRC-Hash-Emo & \checkmark & \checkmark & \checkmark & \checkmark & \checkmark &  &  & \checkmark & \checkmark &  &  &  &  &  &  &  &  &  &  &  &  &  \\ 
$\;\;\;$ NRC-Hash-Sent  &  & \checkmark & \checkmark & \checkmark & \checkmark &  &  & \checkmark &  &  &  &  &  &  &  &  &  &  &  &  &  &  \\ 
$\;\;\;$ NRC-Hashtag-Sent. & \checkmark &  & \checkmark & \checkmark &  &  &  &  &  &  &  &  &  &  &  &  &  &  &  &  &  &  \\ 
$\;\;\;$ NRC10E & \checkmark & \checkmark & \checkmark &  &  &  &  & \checkmark &  &  &  &  &  &  &  &  &  &  &  &  &  &  \\ 
$\;\;\;$ Sentiment140 & \checkmark & \checkmark & \checkmark & \checkmark &  &  &  & \checkmark &  &  &  &  &  &  &  &  &  &  &  &  &  &  \\ 
$\;\;\;$ SentiStrength &  & \checkmark & \checkmark &  &  &  &  & \checkmark &  &  &  &  &  &  &  &  &  &  &  &  &  &  \\ 
$\;\;\;$ SentiWordNet & \checkmark & \checkmark & \checkmark & \checkmark & \checkmark &  &  & \checkmark &  &  &  &  &  &  &  &  &  &  &  &  &  &  \\ 
$\;\;\;$ Vader &  &  &  &  & \checkmark &  &  &  &  &  &  &  &  &  &  &  &  &  &  &  &  &  \\ 
$\;\;\;$ Word.Affect &  &  & \checkmark &  &  &  &  &  &  &  &  &  &  &  &  &  &  &  &  &  &  &  \\ 
$\;\;\;$ In-house lexicon & \checkmark &  &  &  &  &  &  &  &  \checkmark &  &  &  &  &  &  &  & \checkmark &  &  &  &  &  \\ 
Linguistic Features &  &  &  &  &  &  &  &  & \checkmark &  &  &  &  &  &  &  &  &  &  &  &  &  \\ 
$\;\;\;$ Dependency Parser &  &  &  &  &  &  &  &  & \checkmark &  &  &  &  &  &  &  &  &  &  &  &  &  \\ 
\hline
\end{tabular}}
\end{center}
\vspace*{-2mm}
\caption{Feature types (resources) used by the participating systems. Teams are indicated by their rank.}
\vspace*{3mm}
\label{tab:features}
\end{table*}

\begin{table*}[htbp]
\begin{center}
\resizebox{0.9\textwidth}{!}{
\begin{tabular}{lcccccccccccccccccccccc}
\hline
   &\multicolumn{22}{c}{\bf Team}\\
Regression	&1 &2 &3 &4 &5 &6 &7 &8 &9 &$*$ &10 &11 &12 &13  &14 &15 &16 &17 &18 &19 &20 &21\\ \hline
AdaBoost &  &  & \checkmark &  &  &  &  &  &  &  &  &  &  &  &  &  &  &  &  &  &  &  \\ 
Gradient Boosting &  &  & \checkmark & \checkmark &  &  &  &  &  &  &  &  &  &  &  &  & \checkmark &  &  &  &  &  \\ 
Linear Regression &  &  &  & \checkmark &  &  &  &  &  &  &  &  &  &  &  &  &  &  &  &  &  &  \\ 
Logistic Regression &  &  &  &  &  &  &  &  &  & \checkmark &  &  &  &  &  &  &  &  & \checkmark &  &  &  \\ 
Neural Network & \checkmark &  &  & \checkmark &  & \checkmark & \checkmark & \checkmark &  &  &  &  & \checkmark & \checkmark &  & \checkmark &  &  &  & \checkmark & \checkmark & \checkmark \\ 
Random Forest &  & \checkmark & \checkmark &  &  &  &  &  & \checkmark &  &  &  &  &  &  &  &  &  &  &  &  &  \\ 
SVM or SVR &  &  & \checkmark & \checkmark & \checkmark &  &  &  &  &  &  &  & \checkmark &  &  & \checkmark & \checkmark & \checkmark &  & \checkmark &  &  \\ 
Ensemble & \checkmark &  & \checkmark &  &  &  &  &  &  &  &  &  & \checkmark &  &  & \checkmark &  &  &  & \checkmark &  &  \\ 
\hline
\end{tabular}}
\end{center}
\vspace*{-2mm}
\caption{Regression methods used by the participating systems. Teams are indicated by their rank.}
\vspace*{3mm}
\label{tab:regression}
\end{table*}

\begin{table*}[htbp]
\begin{center}
\resizebox{0.9\textwidth}{!}{
\begin{tabular}{lcccccccccccccccccccccc}
\hline
   &\multicolumn{22}{c}{\bf Team}\\
Tools	&1 &2 &3 &4 &5 &6 &7 &8 &9 &$*$ &10 &11 &12 &13  &14 &15 &16 &17 &18 &19 &20 &21\\ \hline
AffectiveTweets-Weka & \checkmark & \checkmark & \checkmark &  &  &  &  & \checkmark &  &  &  &  &  &  &  &  &  &  &  & \checkmark &  &  \\ 
Gensim & \checkmark &  &  & \checkmark &  &  &  &  &  &  &  &  &  &  &  &  &  &  &  &  &  &  \\ 
Glove &  &  &  & \checkmark &  & \checkmark & \checkmark & \checkmark &  &  &  &  &  &  &  &  &  &  &  & \checkmark &  &  \\ 
Keras & \checkmark & \checkmark &  & \checkmark & \checkmark & \checkmark & \checkmark & \checkmark &  &  &  &  & \checkmark &  &  & \checkmark &  &  &  & \checkmark & \checkmark &  \\ 
LIBSVM &  &  &  &  &  &  &  &  &  &  &  &  &  &  &  &  &  & \checkmark &  &  &  &  \\ 
NLTK &  &  &  & \checkmark &  & \checkmark &  &  &  &  &  &  &  &  &  &  &  &  &  &  &  &  \\ 
Pandas & \checkmark &  &  & \checkmark &  &  &  &  &  &  &  &  &  &  &  &  & \checkmark &  &  &  &  &  \\ 
PyTorch &  &  &  &  &  &  &  &  &  &  &  &  &  &  &  &  &  &  &  & \checkmark &  &  \\ 
Sci-kit learn & \checkmark &  & \checkmark & \checkmark &  & \checkmark &  &  &  &  &  &  & \checkmark &  &  & \checkmark & \checkmark &  &  &  &  &  \\ 
TensorFlow & \checkmark &  &  & \checkmark &  &  & \checkmark & \checkmark &  &  &  &  & \checkmark & \checkmark &  &  &  &  &  &  &  &  \\ 
Theano & \checkmark &  &  &  &  &  &  & \checkmark &  & \checkmark &  &  &  &  &  &  &  &  &  &  &  &  \\ 
TweetNLP &  & \checkmark &  &  &  &  &  &  &  &  &  &  &  &  &  &  &  &  &  &  &  &  \\ 
TweeboParser &  &  &  &  &  &  &  &  &  &  &  &  &  & \checkmark &  &  &  &  &  &  &  &  \\ 
Tweetokenize &  &  & \checkmark &  &  &  &  &  &  &  &  &  &  &  &  &  &  &  &  &  &  &  \\ 
Word2Vec & \checkmark & \checkmark &  & \checkmark &  &  &  & \checkmark &  &  &  &  &  &  &  &  &  &  &  &  &  &  \\ 
XGBoost &  &  & \checkmark & \checkmark &  &  &  &  &  &  &  &  &  &  &  &  &  &  &  &  &  &  \\ 
\hline
\end{tabular}}
\end{center}
\vspace*{-2mm}
\caption{Tools and libraries used by the participating systems. Teams are indicated by their rank.}
\label{tab:tools}
\end{table*}

In the subsections below, we briefly summarize the three top-ranking systems. The Appendix (8.3) provides participant-provided summaries about each system.
See system description papers for detailed descriptions.

\subsection{Prayas: Rank 1}
The best performing system, {\it Prayas}, used an ensemble of three different models: 
The first is a feed-forward neural network whose input vector is formed by concatenating the average word embedding vector with the lexicon features vector provided by the AffectiveTweets package \cite{MohammadB17starsem}. These embeddings were trained on a collection of 400 million tweets \cite{godin2015named}.
The network has four hidden layers and uses rectified linear units as activation functions. Dropout is used a regularization mechanisms and the output layer consists of a sigmoid neuron.  
The second model treats the problem as a multi-task learning problem with the labeling of the four emotion intensities as the  four sub-tasks. Authors use the same neural network architecture as in the first model, but the weights of the first two network layers are shared across the four subtasks. The weights of the last two layers are independently optimized for each subtask.
In the third model, the word embeddings of the words in a tweet are concatenated and fed into a deep learning architecture formed by LSTM, CNN, max  pooling, fully connected layers. Several architectures based on these layers are explored.
The final predictions are made by combining the first two models with three variations of the third model into an ensemble. A weighted average of the individual predictions is calculated using cross-validated performances as the relative weights. Experimental results show that the ensemble improves the performance of each individual model by at least two percentage points.

\subsection{IMS: Rank 2}

IMS applies a random forest regression model to a representation formed by concatenating three vectors:
1. a feature vector drawn from existing affect lexicons, 2.  a feature vector drawn from expanded affect lexicons, and 3. the output of a neural network.
The first vector is obtained using the lexicons implemented in the AffectiveTweets package. The second is based on an extended lexicons built from feed-forward neural networks trained on word embeddings. The gold training words are taken from existing affective norms and emotion lexicons: NRC Hashtag Emotion Lexicon \cite{Mohammad2012,COIN:COIN12024}, affective norms from \citet{warriner2013norms}, \citet{brysbaert2014concreteness}, and  ratings for happiness from \citet{dodds2011temporal}. The third vector is taken from the output of neural network that combines CNN and LSTM layers.  

\subsection{SeerNet: Rank 3}
{\it SeerNet} creates an ensemble of various regression algorithms (e.g, SVR, AdaBoost, random forest, gradient boosting). Each regression model is trained on a representation formed by the affect lexicon features (including those provided by AffectiveTweets) and word embeddings. Authors also experiment with different word embeddings models: Glove, Word2Vec, and Emoji embeddings \cite{eisner2016emoji2vec}.

\section{Conclusions}
We conducted  the first shared task on detecting  the intensity of emotion felt by the speaker of a tweet.
We created the 
emotion intensity dataset 
using best--worst scaling and crowdsourcing.
We created a benchmark regression system
and conducted experiments to show that affect lexicons, especially those with fine word--emotion association
scores, are 
useful in determining emotion intensity. 

Twenty-two teams participated in the shared task, with the best system
obtaining a Pearson correlation of 0.747 with the gold annotations on the test set.  
As in many other machine learning competitions, the top ranking systems used ensembles of multiple models (Prayas-rank1, SeerNet-rank3). IMS, which ranked  second, used  random forests, which are ensembles of multiple decision trees.
The top eight systems also made use of a substantially larger number of affect lexicons to generate features than systems that did not perform as well. It is interesting to note that despite using deep learning techniques, training data, and large amounts of unlabeled  data, the best systems are finding it beneficial to include features  drawn  from affect lexicons.

We have  begun work on creating emotion intensity datasets for other emotion categories beyond anger, fear, sadness, and joy. We are also creating a dataset annotated for valence, arousal, and dominance. These annotations will be done for  English, Spanish, and Arabic tweets.  The  datasets will be used in the upcoming SemEval-2018 Task \#1: Affect in Tweets \cite{SemEval2018Affect}.\footnote{http://alt.qcri.org/semeval2018/}

\section*{Acknowledgment}
We thank Svetlana Kiritchenko and Tara Small for helpful discussions. We thank Samuel Larkin for help on collecting tweets.

\bibliography{maxdiff}
\bibliographystyle{acl_natbib}

\newpage

\section{Appendix}

\subsection{Best--Worst Scaling Questionnaire used to Obtain Emotion Intensity Scores}
The BWS questionnaire used for obtaining fear annotations is shown below. 

{
\noindent\makebox[\linewidth]{\rule{0.5\textwidth}{0.4pt}}

\noindent {\bf Degree Of Fear In English Language Tweets}\\[-12pt]

\noindent The scale of fear can range from not fearful at all (zero amount of fear) to extremely fearful. One can often infer the degree of fear felt or expressed by a person from what they say. The goal of this task is to determine this degree of fear. Since it is hard to give a numerical score indicating the degree of fear, we will give you four different tweets and ask you to indicate to us:
\begin{itemize}
\item Which of the four speakers is likely to be the MOST fearful, and\\[-20pt] 
\item Which of the four speakers is likely to be the LEAST fearful. 
\end{itemize}


\noindent {\bf Important Notes}
\begin{itemize}
\item This task is about fear levels of the speaker (and not about the fear of someone else mentioned or spoken to).\\[-20pt]
\item If the answer could be either one of two or more speakers (i.e., they are likely to be equally fearful), then select any one of them as the answer.\\[-20pt]
\item Most importantly, try not to over-think the answer. Let your instinct guide you.\\[-13pt]
\end{itemize}

\noindent {\bf EXAMPLE}\\[-8pt]

\noindent Speaker 1: {\it Don't post my picture on FB \#grrr}\\
Speaker 2: {\it If the teachers are this incompetent, I am afraid what the results will be.}\\
Speaker 3: {\it Results of medical test today \#terrified}\\
Speaker 4: {\it Having to speak in front of so many people is making me nervous.}\\

\noindent Q1. Which of the four speakers is likely to be the MOST fearful?\\ 
-- Multiple choice options: Speaker 1, 2, 3, 4 --\\
\noindent Ans: Speaker 3\\[-8pt]

\noindent Q2. Which of the four speakers is likely to be the LEAST fearful?\\ 
-- Multiple choice options: Speaker 1, 2, 3, 4 --\\
\noindent Ans: Speaker 1\\[-20pt]

\noindent\makebox[\linewidth]{\rule{0.5\textwidth}{0.4pt}}

}
\noindent The questionnaires for other emotions are similar in structure.
In a post-annotation survey, the respondents gave the task high scores for clarity of instruction (4.2/5) despite noting that the task itself requires some non-trivial amount of thought (3.5 out of 5 on ease of task).

\begin{figure*}[t!]
\centering
 \includegraphics[width=5.6in]{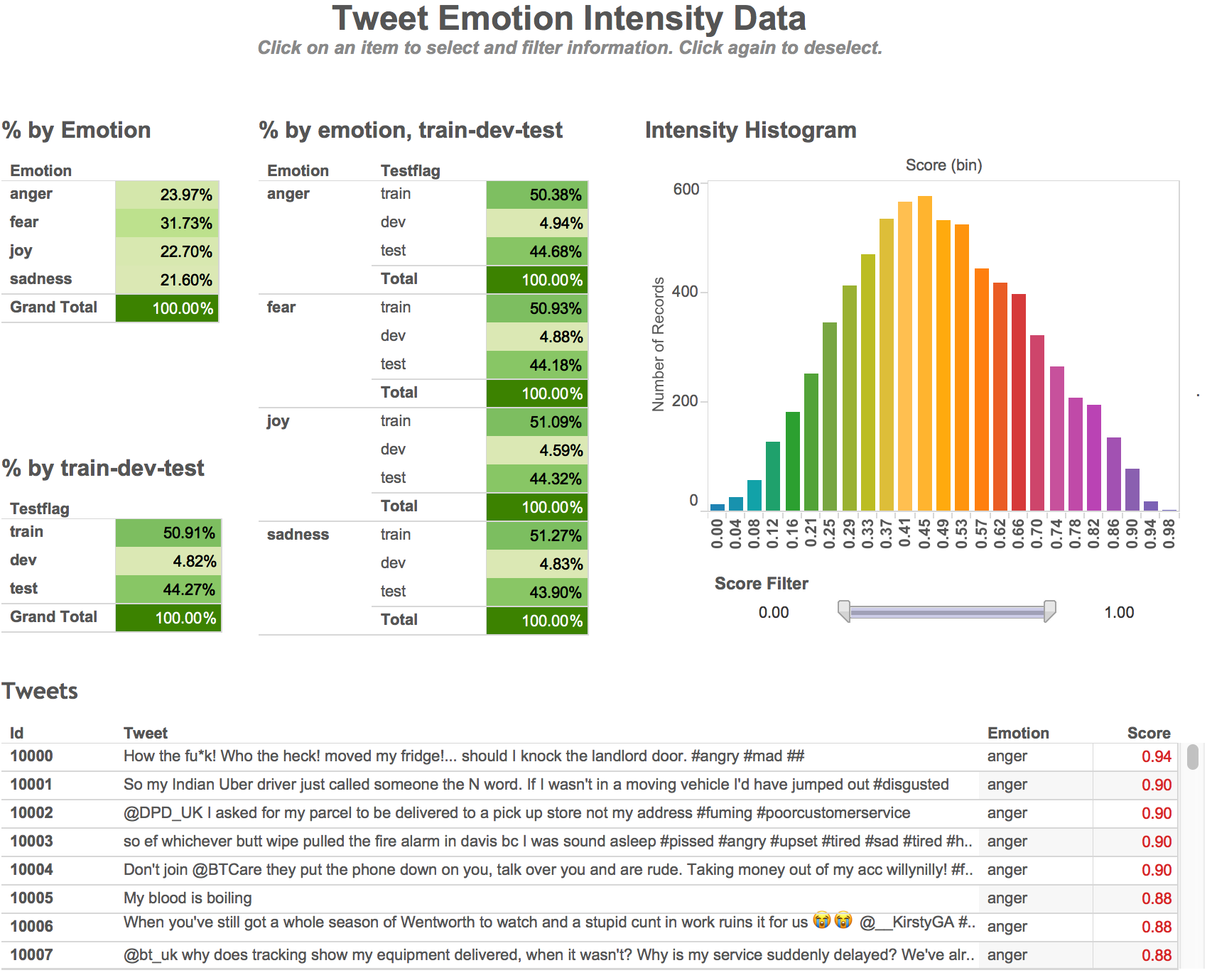}
\caption{Screenshot of the interactive visualization to explore the Tweet Emotion Intensity Dataset.\\
\ed{Available at: http://saifmohammad.com/WebPages/EmotionIntensity-SharedTask.html}}
\label{fig:viz1}
\end{figure*}

\subsection{An Interactive Visualization to Explore the Tweet Emotion Intensity Dataset}

We created an interactive visualization to allow ease of exploration of the {\it Tweet Emotion Intensity Dataset}. This visualization was made public after the the official evaluation period had concluded -- so participants in the shared task did not have access to it when building their system.  It is worth noting that if one intends to evaluate their emotion intensity detection system on the {\it Tweet Emotion Intensity Dataset}, then as a matter of commonly-followed best practices, they should not use the visualization to explore the test data in the system development phase (until all the system parameters are frozen).

The visualization has three main components:\\[-20pt]
\begin{enumerate}
\item Tables showing the percentage of instances in each of
the emotion partitions (train, dev, test). Hovering over a row shows the corresponding number of instances.
Clicking on an emotion filters out data from all other emotions, in all visualization components. Similarly, one can click on just the train, dev, or test
partitions to view information just for that data. Clicking again deselects the item.\\[-22pt]
\item A histogram of emotion intensity scores. A slider that one can use to view only those tweets within a certain score range.\\[-22pt]
\item The list of tweets, emotion label, and emotion intensity scores. \\[-20pt]
\end{enumerate}
\noindent  Notably, the three components are interconnected, such that
 clicking on an item in one component will filter information in all other components to show only the relevant
 details. For example, clicking on `joy' in `a' will cause `b' to show the histogram for only the joy tweets,
 and `c' to show only the `joy' tweets. Similarly one can click on the test/dev/train set, a particular band of
 emotion intensity scores, or a particular tweet.  Clicking again deselects the item. 
One can use filters in combination. For e.g., clicking on fear, test data, and setting the slider for the 0.5 to 1 range,
shows information for only those fear--testdata instances with scores $\ge$ 0.5.

\subsection{AffectiveTweets Weka Package: Implementation Details}

AffectiveTweets includes five filters for converting tweets into feature vectors that can be fed into the large collection of machine learning algorithms implemented within Weka. The package is installed using the \textit{WekaPackageManager} and can be used from the Weka GUI or the command line interface. It uses the \textit{TweetNLP} library \citep{Gimpel11} for tokenization and POS tagging. The filters are described as follows.
\begin{itemize}
\item \textit{TweetToSparseFeatureVector} filter: calculates the following sparse features: word n-grams (adding a NEG prefix to words occurring in negated contexts), 
character n-grams (CN),  POS tags, and  Brown word clusters.\footnote{The scope of negation was determined by a simple heuristic: from the occurrence of a negator word up until a punctuation mark or end of sentence. We used a list of 28 negator words such as {\it no, not, won't} and {\it never}.}
\item \textit{TweetToLexiconFeatureVector} filter:  calculates features from a fixed list of affective lexicons.
\item \textit{TweetToInputLexiconFeatureVector}: calculates features from any lexicon. The input lexicon can have multiple numeric or nominal word--affect associations. 
This filter allows users to experiment with their own lexicons.
\item \textit{TweetToSentiStrengthFeatureVector} filter: calculates positive and negative sentiment intensities for a tweet using the SentiStrength lexicon-based method \cite{ThelwallBP12}
\item \textit{TweetToEmbeddingsFeatureVector} filter: calculates a tweet-level feature representation using pre-trained word embeddings supporting the following aggregation schemes:  average of word embeddings;  addition of word embeddings; and concatenation of the first $k$ word embeddings in the tweet. The package also provides \textit{Word2Vec's} pre-trained word.\footnote{https://code.google.com/archive/p/word2vec/} 
\end{itemize}
\noindent 
Once the feature vectors are created, one can use any of the Weka regression or classification algorithms. Additional filters are under development.
}

\end{document}